\documentclass[a4paper,11pt]{article}
\usepackage{coling2016}

\usepackage{times}
\usepackage{latexsym}

\usepackage{booktabs}

\usepackage{tabularx}
\usepackage{color}
\usepackage{verbatim}
\usepackage{microtype}

\usepackage[utf8]{inputenc}
\usepackage{graphicx}
\usepackage{url}

\usepackage{cleveref}
\newcommand{\com}[1]{}
\usepackage{array}
\usepackage{etoolbox}

\crefname{section}{§}{§§}
\Crefname{section}{§}{§§}
\newcolumntype{R}{>{\raggedright\arraybackslash}X}

\newcolumntype{b}{X}
\newcolumntype{m}{>{\hsize=.43\hsize}R}
\newcolumntype{s}{>{\hsize=.2\hsize}R}
\newcolumntype{n}{>{\hsize=.35\hsize}R}

\title{Survey on the Use of Typological Information \\ in Natural Language Processing}

\author{Helen O'Horan$^{\mathbf{1}}$\thanks{\ \ These authors contributed equally to this work.}, ~ Yevgeni Berzak$^{\mathbf{2}}$\footnotemark[1], ~ Ivan Vuli\'{c}$^{\mathbf{1}}$,  \\
\bf{Roi Reichart}$^{\mathbf{3}}$, ~ {Anna Korhonen}$^{\mathbf{1}}$\\
$^{\mathbf{1}}$ Language Technology Lab, DTAL, University of Cambridge \\
$^{\mathbf{2}}$ CSAIL MIT \\
$^{\mathbf{3}}$ Faculty of Industrial Engineering and Management, Technion, IIT \\
\texttt{helen.ohoran@gmail.com} \hspace{0.5em} \texttt{berzak@mit.edu} \hspace{0.5em} \texttt{iv250@cam.ac.uk} \\ \hspace{0.5em} \texttt{roiri@ie.technion.ac.il} \hspace{0.5em} \texttt{alk23@cam.ac.uk}}

\date{}

\begin{document}

\maketitle

\begin{abstract} 
In recent years linguistic typology, which classifies the world's languages according to their functional and structural properties, has been widely used to support multilingual NLP. While the growing importance of typological information in supporting multilingual tasks has been recognised, no systematic survey of existing typological resources and their use in NLP has been published. This paper provides such a survey as well as discussion which we hope will both inform and inspire future work in the area.
\end{abstract}

%
%

\section{Introduction}
\blfootnote{
    %
    %
    \hspace{-0.65cm}  
    This work is licenced under a Creative Commons Attribution 4.0 International License. License details:
 	\url{http://creativecommons.org/licenses/by/4.0/}
    %
    %
    %
    %
}

One of the biggest research challenges in NLP is the huge global and linguistic disparity in the availability of NLP technology. Still, after decades of research, high quality NLP is only available for a small number of the thousands of languages in the world. Theoretically, we have two solutions to this problem: i) development of universal, language-independent models which are equally applicable to all natural language, regardless of language-specific variation; ii) comprehensive systematisation of all possible variation in different languages.

The field of linguistic typology offers valuable resources for nearing both of these theoretical ideals: it studies and classifies world's languages according to their structural and functional features, with the aim of explaining both the common properties and the structural diversity of languages. Many of the current popular solutions to multilingual NLP: transfer of information from resource-rich to resource-poor languages \cite[inter alia]{Pado-2005,Khapra-2011,das-petrov-2011,Tackstrom-2012}, 
joint multilingual learning \cite[inter alia]{snyder-10,Cohen-2011,Navigli-2012}, 
and development of universal models \cite[inter alia]{Marneffe-2014,nivre-2016}, either assume or explicitly make use of information related to linguistic typology.




While previous work has recognised the role of linguistic typology \cite{Bender-2011}, no systematic survey of typological information resources and their use in NLP to date has been published. Given the growing need for multilingual NLP and the increased use of typological information in recent work, such a survey would be highly valuable in guiding further development. This paper provides such a survey for \textit{structural} typology, the areas of typological theory that consider morphosyntactic and phonological features\footnote{Whilst outside of the scope of this paper, other areas of linguistic typology (e.g., lexico-semantic classifications) are also of significance for the NLP community and should be addressed in future work.}, which has been the main focus of typology research in both linguistics and NLP.

We begin by introducing the field of linguistic typology and the main databases currently available (\cref{sec:overview}). We then discuss the role and potential of typological information in guiding multilingual NLP (\cref{sec:applications}). In \cref{sec:survey} we survey existing NLP work in terms of how typological information has been developed (\ref{sec:survey-development}) and integrated in multilingual application tasks (\ref{sec:survey-applications}). \cref{sec:future} discusses future research avenues, and \cref{sec:commentary-conclusion} concludes.

\section{Overview of Linguistic Typology}
\label{sec:overview}
Languages may share universal features on a deep, abstract level, but the structures found in real-world, surface-level natural language vary significantly. This variation is conventionally characterised into 'languages' (e.g. French, Hindi, Korean)\footnote{Note that there is a lacking consensus on how to define a ‘language’ (as opposed to a dialect, for instance) and the divisions themselves are often arbitrary and/or political. Nonetheless, the divisions are relevant insofar as they are observed in multilingual NLP.}, and linguistic typology describes how these languages resemble or differ from one another. The field comprises three pursuits: the definition of language features and their capacity for variance, the measurement and analysis of feature variance across empirical data, and the explanation of patterns observed in this data analysis. Bickel~\shortcite{Bickel-2007} terms these three pursuits qualitative, quantitative and theoretical typology, respectively.

Typological classifications of languages have strict empirical foundations. These classifications do often support theories of causation, such as historical, areal or phylogenetic relations, but importantly, these hypotheses come second to quantitative data \cite{Bickel-2007}. Indeed, patterns of variance may even run contrary to established theories of relations between languages based on geographical or historical proximity. For instance, Turkish and Korean are typically considered to be highly divergent in lexical features, yet their shared syntactic features make the two languages structurally quite similar. Such indications of similarity are of value for NLP which primarily seeks to model (rather than explain) cross-linguistic variation. 
\begin{table}[!t]
\begin{center}
\def\arraystretch{1.00}
\begin{footnotesize}
\begin{tabularx}{\linewidth}{mnmb}
\toprule
{\bf Name} & {\bf Type} & {\bf Coverage} & {\bf Notes} \\
\midrule
{World Atlas of Language Structures (WALS)} & {Phonology Morphosyntax Lexicosemantics} & {2,676 languages; 192 features; 17\% of features have values} & {Defines language features and provides values for a large set of languages; originally intended for study of areal distribution of features} \\
\midrule
{Syntactic Structures of the World's Languages (SSWL)} & {Morphosyntax} & {262 languages; 148 features; 45\% of features have values} & {Similar to WALS, but differs in being fully open to public editing (Wikipedia-style), and by the addition of numerous example sentences for each feature} \\
\midrule
{Atlas of Pidgin and Creole Language Structures (APiCS)} & {Phonology Morphosyntax Lexicosemantics} & {76 languages;
130 features; 18,526 examples} & {Designed to allow comparison with WALS} \\
\midrule
{PHOIBLE Online} & {Phonology} & {1,672 languages; 2,160 segments} & {Collates and standardises several phonological segmentation databases, in addition to new data} \\
\midrule
{Lyon-Albuquerque Phonological Systems Database (LAPSyD)} & {Phonology} & {422 languages} & {Documents a broader range of features than PHOIBLE, including syllable structures and tone systems; provides bibliographic information and links to recorded samples} \\
\midrule
{URIEL Typological Compendium} & {Phonology Morphosyntax Lexicosemantics} & {8,070 languages and dialects; 284 features; approximately 439,000 feature values} & {Collates features from WALS, SSWL, PHOIBLE, and 'geodata'(e.g. language names, ISO codes, etc.) from sources such as Glottolog and Ethnologue; includes cross-lingual distance measures based on typological features; provides estimates for empty feature values} \\
\bottomrule
\end{tabularx}
\end{footnotesize}
\end{center}
\vspace{-0.0em}
\caption{An overview of major publicly accessible databases of typological information.}
\vspace{-0.1em}
\label{tab:databases}
\end{table}

Typologists define and measure features according to the task at hand. Early studies, focused on word order, simply classified languages as SVO (Subject, Verb, Object), VSO, SOV, and so forth \cite{Greenberg-1963}. There are now more various and fine-grained studies based on a wide range of features, including phonological, semantic, lexical and morphosyntactic properties (see \cite{Bickel-2007,Daniel-2011} for an overview and further references). While a lot of valuable information is contained in these linguistic studies, this information is often not readily usable by NLP due to factors such as information overlap and differing definitions across studies. However, there is also a current trend towards systematically collecting typological information from individual studies in publicly-accessible databases, which are suitable for direct application in NLP (e.g., for defining features and their values). 

Table~\ref{tab:databases} presents a selection of current major databases, including the Syntactic Structures of the World's Languages (SSWL) \cite{sswl}, the World Atlas of Language Structures (WALS) \cite{wals-2013}, the Phonetics Information Base and Lexicon (PHOIBLE) \cite{phoible}, the URIEL Typological Compendium \cite{Littel-2016}, the Atlas of Pidgin and Creole Language Structures (APiCS) \cite{apics}, and the Lyon-Albuquerque Phonological Systems Database (LAPSyD) \cite{Maddieson-2013}. The table provides some basic information about these databases, including type, coverage, and additional notes. From these databases, WALS is currently by far the most commonly-used typological resource in NLP due to its broad coverage of features and languages.

We next discuss the potential of typological information to guide multilingual NLP and the means by which this can be done.

\section{Multilingual NLP and the Role of Typologies}
\label{sec:applications}

The recent explosion of language diversity in electronic texts has made it possible for NLP to move increasingly towards multilingualism. The biggest challenge in this pursuit has been resource scarcity. In order to achieve high quality performance, NLP algorithms have  relied heavily on manually crafted resources such as large linguistically-annotated datasets (treebanks, parallel corpora, etc.) and rich lexical databases (terminologies, dictionaries, etc.). While such resources are available for key NLP tasks (POS tagging, parsing, etc.) in well-researched languages (e.g. English, German, and Chinese), for the majority of other languages they are lacking altogether. Since resource creation is expensive and cannot be realistically carried out for all tasks in all languages, much recent research in multilingual NLP has investigated ways of overcoming the resource problem.

One avenue of research that aims to solve this problem has been unsupervised learning, which exploits unlabelled data that is now available in multiple languages. Over the past two decades increasingly sophisticated unsupervised methods have been developed and applied to a variety of tasks and in some cases also to multiple languages \cite[inter alia]{cohen-smith-2009,Reichart-2010,snyder-10,spitkovsky-2011,Goldwasser-2011,Baker-2014}. However, while purely unsupervised approaches are appealing in side-stepping the resource problem, their relatively low performance has limited their practical usefulness  \cite{Tackstrom-2013}. More success has been gained with solutions that use some form of supervision or guidance to enable NLP for less-resourced languages \cite[inter alia]{Naseem:10,Zhang-2012,Tackstrom-2013}. In what follows, we consider three such solutions: language transfer, joint multilingual learning, and the development of universal models. We discuss the guidance employed in each, paying particular attention to typological guidance. 

\paragraph{Language Transfer}
This very common approach exploits the fact that rich linguistic resources do exist for some languages. The idea is to use them for less-resourced languages via data (i.e. parallel corpora) and/or model transfer. This approach has been explored widely in NLP \cite{hwa-2005,McDonald-2011,Petrov-2012,Zhang-2015}. It has been particularly popular in recent research on dependency parsing, where a variety of methods have been explored. For example, most work for resource-poor languages has combined delexicalised parsing with cross-lingual transfer (e.g. \cite{Zeman-2008,McDonald-2011,sogaard-2011,Rosa:2015acl}). Here, a delexicalised parser is first trained on a resource-rich source language, with both languages POS-tagged using the same tagset, and then applied directly to a resource-poor target language.  

While such a transfer approach outperforms unsupervised learning, it does not achieve optimal performance. One potential reason for this is that the tagset used by a POS tagger may not fit a target language which exhibits significantly different morphological features to a source language for which the tagset was initially developed \cite{Petrov-2012}. Although parallel data can be used to give additional guidance which improves transfer \cite{McDonald-2011}, such data are only available for some language pairs and cannot be used in truly resource-poor situations. 

An alternative direction that has recently emerged uses typological information as a form of non-parallel guidance in transfer. This direction capitalises on the fact that languages do exhibit systematic cross-lingual connections at various levels of linguistic description (e.g. similarities in language structure), despite their great diversity. Captured in typological classifications at the level of generalisation useful for NLP, such information can be used to support multilingual NLP in a variety of ways \cite{Bender-2011}. For example, it can be used to define the similarity between two languages with respect to the linguistic information one hopes to transfer; it can also help to define the optimal degree, level and method of transfer. For example, direct transfer of POS tagging is more likely to succeed when languages are sufficiently similar in terms of morphology in particular \cite{Hana:2004emnlp,Wisniewski:2014emnlp}. 


Typological information has been used to guide language transfer mostly in the areas of part-of-speech tagging and parsing, e.g. \cite{cohen-smith-2009,McDonald-2011,bergkirkpatrick-klein-2010,Naseem-2012,Tackstrom-2013}. Section~\ref{sec:survey} surveys such works in more detail.

\paragraph{Multilingual Joint Learning}
Another approach involves learning information for multiple languages simultaneously, with the idea that the languages will be able to support each other \cite{snyder-10,Navigli-2012}. This can help in the challenging but common scenario where none of the languages involved has adequate resources. This applies even with English, where annotations needed for training basic tools are primarily available only for newspaper texts and a handful of other domains. In some areas of NLP, e.g. word sense disambiguation \cite{Navigli-2012}, multilingual learning has outperformed independent learning even for resource-rich languages, with larger gains achieved by increasing the number of languages.

Success has also been achieved on morphosyntactic tasks. For example, \newcite{snyder-10} observes that cross-lingual variations in linguistic structure correspond to systematic variations in ambiguity, so that what one language leaves implicit, another one will not. For instance, a given word may be tagged as either a verb or a noun, yet its equivalent in other languages may not present such ambiguity. Together with his colleagues, Snyder exploited this variation to improve morphological segmentation, POS tagging, and syntactic parsing for multiple languages. \newcite{Naseem-2012} introduced a selective sharing approach to improve multilingual dependency parsing where the model first chooses syntactic dependents from all the training languages and then selects their language-specific ordering to tie model parameters across related languages. Because the ordering decisions are influenced by languages with similar properties, this cross-lingual sharing is modelled using typological features. In such works, typological information has been used to facilitate the matching of structural features across languages, as well as in the selection of languages between which linguistic information should be shared.

\paragraph{Development of Universal Models}

A long-standing goal that has gained renewed interest recently is the development of language-independent (i.e. {\em universal}) models for NLP \cite{Bender-2011,Petrov-2012}. Much of the recent interest has been driven by the Universal Dependencies (UD) initiative. It aims to develop cross-linguistically consistent treebank annotation for many languages for the purposes of facilitating multilingual parser development and cross-lingual learning \cite{nivre-2016}. The annotation scheme is largely based on universal Stanford dependencies \cite{Marneffe-2014} and universal POS tags \cite{Petrov-2012}. UD treebanks have been developed for 40 languages to date. Whilst still biased towards contemporary Indo-European languages, the collection developed by this initiative is gradually expanding to include additional language families.

The development of a truly universal resource will require taking into account typological variation for optimal coverage. For example, while the current UD scheme allows for language-specific tailoring, in the future, language type-specific tailoring may offer a useful alternative, aligned with the idea of universal modeling \cite{Bender-2011}.

\section{Development and Uses of Typological Information in NLP}
\label{sec:survey}

Given the outlined landscape of multilingual NLP and its relation to structural typology, we now survey existing approaches for obtaining (\ref{sec:survey-development}) and utilizing (\ref{sec:survey-applications}) typological information to support various NLP tasks.

\subsection{Development of Typological Information for NLP}
\label{sec:survey-development}

Typological information has been obtained using two main approaches: i) extraction from manually constructed linguistic resources, such as the databases reviewed in §\ref{sec:overview}; and ii) automatic learning. The two methods have been used independently and in combination, and both are based on the assumption (be it explicit or implicit) that typological relations may be fruitfully used in NLP.

\paragraph{Manual Extraction from Linguistic Resources}
Manually crafted linguistic resources -- in particular the WALS database -- have been the most commonly used sources of typological information in NLP. To date, syntactic parsing \cite{Naseem-2012,Tackstrom-2013,Zhang-2015,Ammar-2016} and POS tagging \cite{Zhang-2012,Zhang-2016} were the predominant areas for integration of structural information from such databases. In the context of these tasks, the most frequently used features related to word ordering according to coarse syntactic categories. Additional areas with emerging research which leverages externally-extracted typological features are phonological modeling \cite{Tsvetkov-2016,Deri-2016} and language learning \cite{Berzak-2015}.

While information obtained from typological databases has been successfully integrated in several NLP tasks, a number of challenges remain. Perhaps the most crucial challenge is the partial nature of the documentation available in manually-constructed resources. For example, WALS currently covers about 17\% of its possible feature values \cite{wals-2013} (see Table \ref{tab:databases} for feature coverage of other typological databases). The integration of information from different databases is challenging due to differences in feature taxonomies as well as information overlap across repositories. Furthermore, available typological classifications contain different feature types, including nominal, ordinal and interval variables, and features that mix several types of values. This property hinders systematic and efficient encoding of such features in NLP models -- a problem which thus far has only received a partial solution in the form of feature binarisation \cite{Georgi-2010}. Further, typological databases are constructed manually using limited resources, and do not contain information on the distribution of feature values within a given language. This results in incomplete feature characterisations, as well as inaccurate generalisations. For example, WALS encodes only the dominant noun-adjective ordering for French, although in some cases this language also permits the adjective-noun ordering.

Other aspects of typological databases may require feature pruning and preprocessing prior to use. For example, some features in WALS, such as feature 81B ``Languages with two Dominant Orders of Subject, Object, and Verb'' are applicable only to a subset of the world's languages. Currently, no explicit specification for feature applicability is present in WALS or other typological resources. Furthermore, distinct features may encode overlapping information, as in the case of WALS features 81A ``Order of Subject Verb and Object'' and 83A ``Order of Verb and Object'', where the latter can be deduced from the former. Although many of these issues have been noted in previous research \cite{Ostling-2015}, there are currently no standard procedures for preprocessing typological databases for NLP use. 

Despite the caveats presented above, typological resources do offer an abundance of valuable structural information which can be integrated in many NLP tasks. This information is currently substantially underutilised. Out of 192 available features in WALS, only a handful of word order features are typically used to enhance multilingual NLP. Meanwhile, the complementary information on additional languages and feature types offered by other repositories has, to our knowledge, rarely been exploited in NLP. This readily-available information could be used more extensively in tasks such as POS tagging and syntactic parsing, which have already gained from typological knowledge, and it could also be used to support additional areas of NLP. 

\paragraph{Automatic Learning of Typological Information}
\label{automatic-learning}
The partial coverage of existing typological resources, stemming from the difficulty of obtaining such information manually, have sparked a line of work on automatic acquisition of typological information. Here too, WALS has been the most common reference for defining the features to be learned.   

Several approaches were introduced for automatic induction of typological information through multilingual word alignments in parallel texts. \newcite{Mayer-2012} use alignments to induce language similarities, and use this approach to support learning of fine-grained features, such as the typology of person interrogatives (e.g., English "who"). In \newcite{Ostling-2015} multilingual word alignments are used to project POS tags and syntactic trees for translations of the New Testament, and subsequently learn typological information relating to word order. The predicted typological features, when evaluated against WALS, achieve high accuracy. This method not only extends WALS word order documentation to hundreds of new languages, but also quantifies the frequency of different word orders across languages -- information that is not available in manually crafted typological repositories.

Typological information can also be extracted from Interlinear Glossed Text (IGT). Such resources contain morphological segmentation, glosses and English translations of example sentences collected by field linguists. \newcite{Lewis-2008} and \newcite{Bender-2013} demonstrate that IGT can be used to extract typological information relating to word order, case systems and determiners for a variety of languages.

Another line of work seeks to increase the coverage of typological information using existing information in typological databases. \newcite{Daume-2007} and \newcite{Bakker-2008} use existing WALS features to learn typological implications of the kind pioneered by Greenberg \shortcite{Greenberg-1963}. Such rules can then be used to predict unknown feature values for new languages. \newcite{Georgi-2010} use documented WALS features to cluster languages, and subsequently predict new feature values using nearest-neighbour projection. A classifier-based approach for predicting new feature values from documented WALS information is presented in \cite{Takamura-2016}. \newcite{Coke-2016} predict word order typological features by combining documented typological and genealogical features with the multilingual alignment approach discussed above. 

An alternative approach for learning typological information uses English as a Second Language (ESL) texts \cite{Berzak-2014}. This work demonstrates that morphosyntactic typological similarities between languages are largely preserved in second language structural usage. It leverages this observation to approximate typological similarities between languages directly from ESL usage patterns and further utilise these similarities for nearest neighbor prediction of typological features. The method evaluates competitively compared to baselines in the spirit of \cite{Georgi-2010} which rely on existing typological documentation of the target language for determining its nearest neighbors.

In addition, a number of studies learned typological information tailored to the particular task and data at hand (i.e. {\em task-based development}). For example,  \newcite{Song-Xia-2014} process Ancient Chinese using Modern Chinese parsing resources. They manually identify and address statistical patterns in variation between monolingual corpora in each language, and ultimately optimise the model performance by selectively using only the Modern Chinese features which correspond to Ancient Chinese features.

Although automatically-learned typological classifications have not been used frequently to date, they hold great promise for extending the use of typological information in NLP. Furthermore, such work offers an additional axis of interaction between linguistic typology and NLP, namely using computational modeling in general and NLP in particular to assist linguistic documentation and analysis of typological information. We discuss the future prospects of these research directions in \cref{sec:commentary-conclusion}.

\subsection{Uses of Typological Information in NLP}
\label{sec:survey-applications}

\paragraph{Multilingual Syntactic Parsing}

As mentioned in \cref{sec:survey-development}, the main area of NLP in which information from structural typology has been exploited thus far is multilingual dependency parsing. In this task, a priori information about the predominant orderings of syntactic categories across languages are used to guide models when parsing a resource-poor language and using training data from other languages. This information is available in typological resources (e.g., WALS) which, among a variety of other syntactic features, list the dominant word orderings for many languages (see Table~\ref{tab:databases}). 

A seminal work that integrates typological word order information in multilingual dependency parsing \cite{Naseem-2012} presents the idea of ``selective sharing'' between source and target languages. In brief, while the identity of possible dependents for a given syntactic category is (hypothesised to be) language-universal, their ordering is language-specific. The work then presents a generative multilingual parsing model in which dependent ordering parameters are conditioned on word order typology, obtained from WALS. Specifically, the paper utilises the following word order features (henceforth WALS Basic word Order, WBO): 81A (Subject Verb and Object), 85A (Adposition and Noun), 86A (Genitive and Noun), 87A (Adjective and Noun), 88A (Demonstrative and Noun) and 89A (Numeral and Noun). This information enables the model to take into account dependent orderings only when the source language has a similar word order typology to the target language. In a similar vain, \newcite{Tackstrom-2013} present an instance of the typologically guided selective sharing idea within a discriminative parsing framework. They group the model features into features that encode arc directionality and word order, and those that do not. The former group is then coupled with the same WBO features used by \newcite{Naseem-2012} via feature templates that match the WALS properties with their corresponding POS tags. Additional features that group languages according to combinations of WALS features as well as coarse language groups (Indo-European versus Altaic), result in further improvements in parsing performance. 

\newcite{Zhang-2015} extended the selective sharing approach for discriminative parsing to tensor-based models using the same WBO features as in \cite{Naseem-2012} and \cite{Tackstrom-2013}. While traditional tensor-based parsers represent and assign non-zero weights to all possible combinations of atomic features, this work presents a hierarchical architecture that enables discarding chosen feature combinations. This allows the model to integrate prior typological knowledge, while ignoring uninformative combinations of typological and dependency features. At the same time, it capitalises on the automatisation of feature construction inherent to tensor models to generate combinations of informative typology-based features, further enhancing the added value of typological priors. 

Another successful integration of externally-defined typological information in parsing is the work of \newcite{Ammar-2016}. They present a multilingual parser trained on a concatenation of syntactic treebanks of multiple languages. To reduce the adverse impact of contradicting syntactic information in treebanks of typologically distinct languages, while still maintaining the benefits of additional training data for cross-linguistically consistent syntactic patterns, the parser encodes a language-specific bias for each given input language. This bias is based on the identity of the language and its WBO features as used in \cite{Naseem-2012,Tackstrom-2013,Zhang-2015}. Differently from prior work, their parsing model also encodes all other features in the WALS profile of the relevant language. Overall, this strategy leads to improved parsing performance compared to monolingually trained baseline parsers.

While the papers surveyed above use prior information about word order typology extracted from WALS, word order information for guiding multilingual parsing can also be extracted in a bottom-up, data-driven fashion, without explicit reference to typological taxonomies. For example, in \newcite{sogaard-2011}, training sentences in a source language are selected based on the perplexity of their coarse POS tag sequence under a target language POS language model. This approach essentially chooses sentences that exhibit similar word orderings in both source and target languages, thus realizing a bottom-up variant of the typology-based selective sharing methods discussed above.

There are also several methods which have made use of less explicit typological information. For instance, \newcite{bergkirkpatrick-klein-2010} selectively combine languages in their method for cross-lingual dependency grammar induction using a phylogeny tree, which has been constructed from external (unspecified) knowledge of language families. \newcite{Zeman-2008} demonstrate improved performance of cross-lingually transferred dependency parsers within sets of typologically similar languages (e.g. Swedish-Danish, Hindi-Urdu); they do not explain how languages may be determined as ``closely-related'', though presumably this decision was based on the intuition of the researchers or on widely-acknowledged generalisations.

\paragraph{POS Tagging, Phonological Modeling and Language Learning}
Besides dependency parsing, several other areas have started integrating typological information in various forms. A number of such works revolve around the task of POS tagging. For example, in 
\newcite{Zhang-2012}, the previously discussed WBO features were used to inform mappings from language-specific to a universal POS tagset. In \cite{Zhang-2016}, WBO feature values are used to evaluate the quality of a multilingual POS tagger. 


Another application area which benefited from integration of typological knowledge are phonological models of text. In \cite{Tsvetkov-2016} a multilingual neural phoneme-based language model is trained on several languages using a shared phonological inventory. The model is conditioned on the identity of the language at hand, as well as its phonological features obtained from a concatenation of phonological features from WALS, PHOIBLE and Ethnologue, extracted from URIEL. The resulting model subsumes and outperforms monolingually trained models for phone sequence prediction. \newcite{Deri-2016} use URIEL to obtain phone and language similarity metrics, which are used for adjusting Grapheme to Phoneme (G2P) models from resource rich to resource poor languages. 

\newcite{Berzak-2015} use typological classifications to study language learning. Formalizing the theory of ``Contrastive Analysis'' which aims to analyse learning difficulties in a foreign language by comparing native and foreign language structures, they build a regression model that predicts language-specific grammatical error distributions by comparing typological features in the native and foreign languages.

\section{Typological Information and NLP: What's Next?}
\label{sec:future}
\cref{sec:survey-applications} surveyed the current uses of typological information in NLP. Here we discuss several future research avenues that might benefit from tighter integration of linguistic typologies and multilingual NLP.

\paragraph{Encoding Typological Information in Traditional Machine Learning-based NLP}

One of the major open challenges for typologically-driven NLP is the construction of principled mechanisms for the integration of typological knowledge in machine learning-based algorithms. Here, we briefly discuss a few traditional machine learning frameworks which support encoding of expert information, and as such hold promise for integrating typological information in NLP.

Encoding typological knowledge into machine learning requires mechanisms that can bias {\it learning (parameter estimation)} and {\it inference (prediction)} of the model towards predefined knowledge. Algorithms such as the structured perceptron \cite{Collins:02} and structured SVM \cite{Taskar:04} iterate between an inference step and a parameter update step with respect to gold training labels. 
The inference step is a natural place for encoding external knowledge through constraints. It biases the prediction of the model to agree with external knowledge, which, in turn, affects both the training process and the final model prediction. As typological information often reflects tendencies rather than strict rules, {\it soft constraints} are helpful. Ultimately, an efficient mechanism for {encoding soft constraints into the inference step} is needed.
Indeed, several modeling approaches have been proposed that do exactly this: constraint-driven learning (CODL) \cite{Chang:07}, posterior regularisation (PR) \cite{Ganchev:10}, generalized expectation (GE) \cite{Mann:08}, and dual decomposition \cite{Globerson:08}, among others.
Such approaches have been applied successfully to various NLP tasks where external knowledge is available. 
Examples include POS tagging and parsing \cite{Rush:10,Rush:12}, information extraction \cite{Riedel-2011,Reichart-2012b}, and discourse analysis \cite{Guo:13}, among others. 
In addition to further extensions to the modeling approaches surveyed in §\ref{sec:survey-applications}, these type of frameworks could expedite principled integration of typological information in NLP.

\paragraph{Typologies and Multilingual Representation Learning}
While the traditional models surveyed above assume a predefined feature representation and focus on generating the best prediction of the output labels, a large body of recent NLP research has focused on learning dense real-valued vector representations —- i.e., word embeddings (WEs). WEs serve as pivotal features in a range of downstream NLP tasks such as parsing, named entity recognition, and POS tagging \cite{Turian:2010acl,Collobert:2011jmlr,Chen:2014emnlp}. The extensions of WE models in bilingual and multilingual settings \cite[inter alia]{Klementiev:2012coling,Hermann:2014acl,Coulmance:2015emnlp,Vulic:2016jair} abstract over language-specific features and attempt to represent words from several languages in a language-agnostic manner such that similar words (regardless of the actual language) obtain similar representations. Such multilingual WEs facilitate cross-lingual learning, information retrieval and knowledge transfer. The extent to which multilingual WEs capture word meaning across languages has been recently evaluated in \cite{Leviant-2015} with the conclusion that multilingual training usually improves the alignment between the induced WEs and the meaning of the participating words in each of the involved languages.

Naturally, as these models become more established and better understood, the challenge of external knowledge encoding becomes more prominent. Recent work has examined the ability to map from word embeddings to interpretable typological representations \cite{Qian-2016}. Furthermore, a number of works \cite{Faruqui:2015naacl,Rothe:2015acl,osborne-16,mrksic:2016:naacl} proposed means through which external knowledge from structured knowledge bases and specialised linguistic resources can be encoded in these models. The success of these works suggests that more extensive integration of external linguistic knowledge in general, and typological knowledge in particular, is likely to play a key role in the future development of WE representations.

\paragraph{Can NLP Support Typology Construction?}
As discussed in §\ref{sec:survey}, typological resources are commonly constructed manually by linguists. Despite the progress made in recent years in the digitisation and collection of typological knowledge in centralised repositories, their coverage remains limited. Following the work surveyed in §\ref{automatic-learning} on automatic learning of typological information, we believe that NLP could play a much larger role in the study of linguistic typology and the expansion of such resources. Future work in these directions will not only assist in the global efforts for language documentation, but also substentially extend the usability of such resources for NLP purposes.

\section{Commentary; conclusion}
\label{sec:commentary-conclusion}


This paper has provided a survey of linguistic typologies and the many recent works in multilingual NLP that have benefited from such resources. We have shown how combined knowledge of linguistic universals and typological variation has been used to improve NLP by enabling the use of cross-linguistic data in the development and application of resources. Promising examples of both explicit and implicit typological awareness in NLP have been presented. We have concluded with a discussion on how typological information could be used to inform improved experimental and conceptual practice in NLP. We hope that this survey will be useful in both informing and inspiring future work on linguistic typologies and multilingual NLP.

\section*{Acknowledgments}
This work is supported by ERC Consolidator Grant LEXICAL (no 648909) and by the Center for Brains, Minds and Machines (CBMM) funded by the NSF STC award CCF-1231216. The authors are grateful to the anonymous reviewers for their helpful comments and suggestions.

\bibliography{coling2016}
\bibliographystyle{acl}

\end{document}